# Practical Selection of SVM Supervised Parameters with Different Feature Representations for Vowel Recognition

[1]Rimah Amami, [2]Dorra Ben Ayed, [3]Noureddine Ellouze
*[1], Corresponding Author* rimah.amami@yahoo.fr
[1,2,3] Department of Electrical Engineering, National School of Engineering of Tunis
Université de Tunis El Manar, Tunisia

## Abstract

*It is known that the classification performance of Support Vector Machine (SVM) can be conveniently affected by the different parameters of the kernel tricks and the regularization parameter, C. Thus, in this article, we propose a study in order to find the suitable kernel with which SVM may achieve good generalization performance as well as the parameters to use. We need to analyze the behavior of the SVM classifier when these parameters take very small or very large values. The study is conducted for a multi-class vowel recognition using the TIMIT corpus. Furthermore, for the experiments, we used different feature representations such as MFCC and PLP. Finally, a comparative study was done to point out the impact of the choice of the parameters, kernel trick and feature representations on the performance of the SVM classifier*

## 1. Introduction

Support Vector Machine (SVM) has been a very successful supervised algorithm for solving two-class and multi-class recognition problems. SVM is a method which is based on the Vapnic-Chervonenkis (VC) theory and the principle of structural risk minimization (SRM) [1][2]. Furthermore, the excellent learning performance of this method comes out from the fact that SVM apply a linear algorithm to the data in a high dimensional space.

The application of SVM to the automatic speech recognition (ASR) problem requires solving three main problems: First, the variable time duration of each utterance and features used to their representation. Second, the size of the databases used in speech recognition. The last big problem is related to the choice of parameters which SVM and kernel tricks needs such as the kernel width Gamma ($\sigma$) and the regularization parameter C which is an unintuitive parameter.

The performance is, indeed, depending on the parameters of both SVM and kernel function which aims to help SVM to get an optimal separating hyperplan in the feature space. On the other hand, the choice of the feature which represent vowel samples play a key role as well as SVM. Thus, in this paper, we focus, also, on the feature representations to use.

First, we have selected the most two popular features representations in this research area such as MFCC and PLP. We have also compared the performance of each two different strategies of extracting frames by adjusting the number of the different frames used to see how the ASR system behaves when we increase the number of feature to use.

Furthermore, in this paper, we try to answer the following question: when the samples are mapped nonlinearly into a high dimensional feature space, what is the suitable kernel trick? When one of the kernel tricks is selected, how to control their complexity to avoid overfitting? What is the suitable value of the regularization parameter C?
All of these have to be carefully turned in practical use of SVM in an automatic phoneme recognition filed.

The remainder of this paper is organized as follows: In the next section, a short introduction to Support Vector Machines will be given. In section 3 some kernel functions will be presented. In section 4, a brief overview of feature extraction and the feature representation used is presented. In sections 5 and 6, experiments are made to see how these kernels and SVM perform on real-world data.

## 2. Support vector machine

Support Vector Machine (SVM) is a learning machine which was developed by Vladimir Vapnik in aims to construct decision functions in the input space based on the theory of Structural Risk Minimization [3][4].

Furthermore, SVM consists of constructing one or several hyperplans in order to separate the different classes. Nevertheless, an optimal hyperplan must be found. Vapnik and Cortes [5] defined an optimal hyperplan as the linear decision function with maximal margin between the vectors of the two classes. We consider the optimal hyperplan if it is separated the examples without error and if the distance between the closest example and the hyperplan is maximal. The hyperplan can be described as:

$$W^T x + b = 0, x \in R^d \qquad (1)$$

The samples closed to the hyperplan boundaries are called « support vectors ». They are used to decide which hyperplan should be selected since this set of vectors is separated by the optimal hyperplan. SVM is basically used as a linear decision function when the data are separable. However, in this paper, we consider that the data are linearly non-separable. Therefore, we should introduce a nonlinear function with a nonnegative variables ($\varphi(\xi_i)$) which can map the data in a high-dimensional feature space where they are linearly separable. The optimal hyperplan in a nonlinear space can be determined by the vector W which minimizes the functional:

$$\varphi(W, \xi) = \frac{1}{2} \|W\|^2 + C \sum_{i=1}^{l} \xi_i \qquad (2)$$

Where $\xi$ is a slack variable and C a pre-specified value which is used to control the amount of regularization.

Furthermore, it must be pointed out that, in this work, a multi-class recognition problem is decoupling to a two-class problem [7]. Therefore, we used the one-against-one approach proposed by Knerr et al. [8]. This approach consists of constructed k (k - 1)/2 classifiers where each one trains samples from two classes. For the recognition decision making, the majority voting strategy was applied.

## 3. Kernel functions

The kernel functions are one of the major tricks of SVM. Those functions are used when the samples are linearly non-separable. Thus, the kernel tricks extends the class of decision functions to the non-linear case by mapping the samples from the input space X into a high-dimensional feature R without ever having to compute the mapping explicitly, in the hope that the samples will gain meaningful linear structure in R.

Furthermore, the kernel function can be interpreted as a measure of similarity between the samples $x_i$ and $x_j$ [9] which it allows SVM classifiers to perform separations even with very complex boundaries. There are several possibilities for the choice of this kernel function, including polynomial, sigmoid, RBF. In the sequel of this paper, we will try to find the best choice of the kernels function. Below, there is a brief overview of some kernel functions available from the existing literature.

### 3.1. Polynomial kernel

The Polynomial kernel is a non-stationary kernel. It is well suited for problems where all the training samples are normalized. The parameters which must be settled are the slope gamma. The constant term *r* and the polynomial degree *d* (hence *d*=3, *r*=0).

$$K(x_i, x_j) = (\sigma x_i^T x_j + r)^d, \sigma > 0 \qquad (3)$$

### 3.2. RBF kernel

RBF (Gaussian) kernels are a family of kernels where a distance measure is smoothed by a radial function (exponential function) [10]. This kernel nonlinearly maps samples into a higher dimensional space so it, unlike the linear kernel, can handle the case when the relation between class labels and attributes is nonlinear.

Furthermore, the linear kernel is a special case of RBF [11] since the linear kernel with a penalty parameter C has the same performance as the RBF kernel with some parameters (C, Gamma).

$$K(x_i, x_j) = \exp(-\sigma \|x_i - x_j\|^2), \sigma > 0 \qquad (4)$$

The adjustable parameter $\sigma$ plays a major role in the performance of the kernel, and should be carefully tuned. If overestimated, the exponential will behave almost linearly and the higher-dimensional projection will start to lose its non-linear power. In the other hand, if underestimated, the function will lack regularization and the decision boundary will be highly sensitive to noise in training data. Thus, the behavior of SVM depends on the choice of the width parameter $\sigma$.

### 3.3. Sigmoid kernel

The kernel must satisfy Mercer's theorem, and that requires that the kernel be positive definite. However, the Sigmoid kernel, which, despite its wide use, it is not positive semi-definite for certain values of its parameters. Thus, the parameters $\sigma$, $r$ must be properly chosen otherwise, the results may be drastically wrong, so much so that the SVM performs worse than random.

$$K(x_i, x_j) = \tanh(\sigma x_i^T x_j + r) \qquad (5)$$

We can view $\sigma$ as a scaling parameter of the input samples, and $r$ as a shifting parameter that controls the threshold of mapping (hence $r = 0$). In general, the sigmoid kernel is not better than RBF and linear kernels [11].

## 4. Feature extraction

The vowel recognition task can be roughly divided into two stages: The feature extraction and recognition. One of the fundamental denominators of all recognition system is the feature extraction since all of the information necessary to distinguish vowel is preserved during this stage [13]. So, if important information are lost during the feature extraction, the performance of the following recognition stage will be inherently deteriorated which will affect the system [14][15].

Feature extraction can be considered as a way to reduce the dimensionality of the input data, a reduction which leads necessarily to lose some information. In other words, the vowel signals will be segmented into frames and extract features from each frame. Furthermore, we used a filter with pre-emphasis factor time domain equal to 0.95 in aims to remove unwanted frequency components from the signal. Afterwards, the vowel signals were segmented into frames of 16 milliseconds length with optional overlap of 1/3~1/2 of the frame size. Thus, if the sample rate is 16 kHz and the frame size is 256 sample points, then the frame duration is 256/16000 = 0.016 sec = 16 ms.

Additional, if the overlap is 128 points, then the frame rate is 16000/(256-128) = 125 frames per second. Moreover, for the FFT-based features, each frame was weighted by a Hamming window in order to keep the continuity of the first and the last points in the frame [17]. Due to the impact of feature representations used, in this paper, we tried to find the suitable dimensional feature vectors which generates the best recognition rates (12 or 36-dimensional feature vectors).

In this paper, we are interested by three speech parameterization techniques: MFCC and PLP presented below[18][19].

### 4.1. MFCC

Mel-frequency cepstral coefficients (MFCC) are derived from a type of cepstral representation. Davis and Mermelstein [20] were the first who introduced in 80's the MFCC concept for automatic speech recognition. The main idea of this algorithm consider that the MFCC are the cepstral coefficients calculated from the mel-frequency warped Fourier transform representation of the log magnitude spectrum.

The Delta and the Delta-Delta cepstral coefficients are an estimate of the time derivative of the MFCCs. In order to improve the performance of speech recognition system, an improved representation of speech spectrum can be obtained by extending the analysis to include the temporal cepstral derivative; both first (delta) and second (delta-delta) derivatives are applied [20]. Those coefficients have shown a determinant capability to capture the transitional characteristics of the speech signal that can contribute to ameliorate the recognition task.

### 4.2. PLP

Perceptual Linear Prediction (PLP) is a hybrid of DFT and LP (linear predictive) techniques proposed by Hynek Hermansky. PLP algorithm is based on the short-term spectrum of speech. It modifies the short-term spectrum of the speech by several psychophysically based transformations.

Later studies [20][19] have shown that the PLP features outperform MFCC in specific conditions, but generally no large gap in performance was observed between them.

## 5. Experimental setup

In this paper, the speech parameterizations techniques were evaluated on the dialect region DR1from TIMIT speech recognition corpus [21][22].This English dialect is subdivided into training and testing samples. In this study, only the vowel phonemes were utilized {/aa/ ,/ae/ , /ah/, /ao/, /aw/, /ax/, /ax-h/, /axr/, /ay/, /eh/, /er/, /ey/, /ih/, /ix/, /iy/, /ow/, /oy/, /uh/, /uw/, /ux/}[23] [24].

For each vowel phoneme sample, we will extract a K-dimensional feature vectors. Then, we will investigate each vector in order to select only the frames which will be used in the recognition stage.

Next, the vectors will be divided by a norm of the vector. In fact, scaling is a very important optional pre-processing step which leads to avoid the problem occurring when attributes in greater numeric ranges dominating those in smaller numeric ranges. Besides, scaling data may help to avoid numerical difficulties during the calculation as the kernel values usually depend on the inner products of feature vectors, such as the polynomial kernel. So, large attribute values might cause numerical problems.

In the neural network literature [12], this step refers to scaling by the minimum and range of the vector, to make all the elements lay between 0 and 1 or between -1 and 1. Thus in this study, we scale each attribute to the range [0, 1]. Afterwards, for the recognition stage, we utilized for the validation test dataset.

## 6. Experimental results

### 6.1. Experimental results of frames selection

At this stage of experiments, we investigate to find the suitable number of frames to use. Since the middle frames are known that they may contain the most important information about the speech signal, this choice could be considered important.

It must be pointed out that we used two different methods to research the suitable number of frames to utilize: Middle frames and Fuzzy c-means clustering (FCM).

FCM, developed by Dunn in 1973 and improved by Bezdek in 1981, is a method of clustering which allows one set of data to belong to two or more clusters. At this stage of the study, we supposed that we used MFCC 36-dimensional feature vectors, the kernel trick used is RBF and we set C=10, $\sigma$ =0.027.

Table 1. % accuracy with K-frames using FCM and Middle frame methods

| Kernel | K=3 | | K=5 | | K=7 | |
|---|---|---|---|---|---|---|
| | FCM | Middle Frame | FCM | Middle Frame | FCM | Middle Frame |
| Polynomial | 39.08 | 46 | 39.67 | 48.90 | 41.44 | 45.33 |
| RBF | 42.47 | 51.60 | 41.46 | 46.11 | 40.66 | 44.96 |
| Sigmoid | 41.73 | 50 | 41.38 | 49.26 | 41.77 | 45.90 |

As seen in the table 1, 3-middle frames generate the best accuracy with a recognition rate equal to 51.60%. We notice, also, that in our case the performance of the method based on middle frames outreach the performance of FCM method.
An interesting observation is that when we enlarge the number of middle frames, the accuracy decrease (3-MF=51% vs. 7-MF= 45%).
For the next experiments, we will retain the method which present the best recognition rate which is 3-Middle Frames.

### 6.2. Recognition results

It is important to see the behavior of SVM when we include unknown testing data and with different parameters values. A few key questions is to know: How the recognition system will behave with large value of the parameter C? IS MFCC the suitable feature representation to utilize? Must we use the optimum value of $\sigma =2$ found in the grid search using a cross-validation, or investigate more for the optimum values?
More generally, we wanted to observe the behavior of SVM when the parameters take very small or very large values.
At this stage of experiments, our work consists in finding the optimal value of two parameters: The kernel width ($\sigma$) and the regularization parameter C. First, we set the initial values of $\sigma =2$ and C= [10, 100, 1000, 10000] in aim to find the suitable value of C giving the optimal recognition rates.
The experiments results in table 2 showed that SVM-RBF performs very well on different feature representations.

Table 2. % Comparison of recognition accuracy with different kernel functions and feature representations

| C | 10000 | | 1000 | | 100 | | 10 | |
|---|---|---|---|---|---|---|---|---|
| | MFCC | PLP | MFCC | PLP | MFCC | PLP | MFCC | PLP |
| SVM-Polynomial | 40.10 | 38.35 | 40.10 | 38.35 | 40.10 | 38.35 | 40.35 | 38.49 |
| SVM-RBF | 44.44 | 44.56 | 44.44 | 44.56 | 44.29 | 44.56 | 46.00 | 45.80 |
| SVM-Sigmoid | 15.86 | 15.86 | 15.86 | 15.86 | 15.86 | 15.86 | 15.86 | 15.86 |

Note that for the SVM-polynomial, the accuracy went down by range within 2% to 10% compared to SVM-RBF whereas for the sigmoid kernel, the accuracy went down by range within 2% to 35% compared to SVM-RBF.
On the other side, the runtime (training and testing) of the SVM-RBF kernel was better than with SVM-polynomial and SVM-sigmoid. The runtime of SVM-polynomial tend to be relatively long compared to SVM-RBF (72s Vs 630s).
Furthermore, we observed, also, that when value of the penalty parameter C tend to be high, the runtime tend to be more important (i.e. with PLP, SVM-polynomial: c=10000, runtime=407876s Vs C=10, runtime=337). In general, even the runtime of SVM with PLP representation was important than with MFCC representation.
All in all, the penalty C=10 and SVM-RBF with 36-dimensional MFCC feature vectors shows the best performance (acc = 46%) in term of recognition rates and runtime. The SVM-polynomial performs worse (acc = 40.35%), but still with good results and reasonable runtime when C is slow. The SVM-sigmoid comes third (acc =15.86%) and show the worst performance.

Based on figure 1, we notice that the recognition rates were considerably improves with smaller $\sigma$. We observed that with a small value of gamma, the accuracy tend to be better (51% Vs 44%).

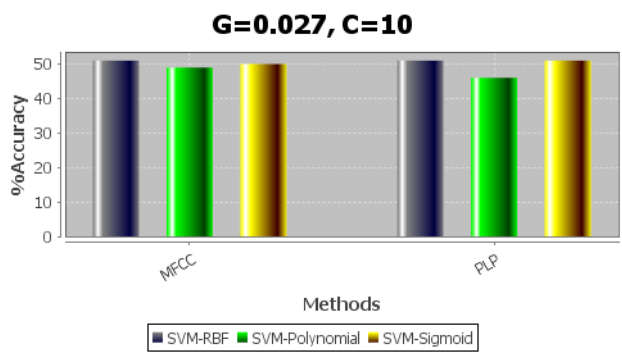

It must be pointed out that even the SVM-sigmoid which gives the worst results in the previous experiments outreach the SVM-polynomial. Meanwhile, the PLP representation generates recognition rates slightly worse than MFCC representation.

The result is quite consistent with what we could expect in such a situation: when the parameters are smaller and with a higher dimension feature vector, the system tends to give a better recognition rates.

Meanwhile, choosing the most appropriate kernel highly depends on the recognition problem and searching its parameters can easily become a tedious and complex task.

## 7. Conclusion

The paper has presented different SVM kernels that can be utilized for vowel recognition using different feature representations. We proposed this study in aims to investigate the optimal supervised parameters and features representations.

SVM-RBF with 36-dimensional MFCC feature vectors seems, in our case, to perform better than the different kernels and features representations studied. Moreover, based on our experiments, the kernel width parameter and the penalty parameter tends to be smaller, the accuracy of the vowel system and the runtime improves.

Otherwise, how to efficiently find out which kernel is optimal for a given learning task is still an unsolved problem and hence lays the biggest limitation of SVM. This key question stills a research problem and a cumbersome task.